\documentclass[conference]{IEEEtran}
\IEEEoverridecommandlockouts
\usepackage{cite}
\usepackage{amsmath,amssymb,amsfonts}
\usepackage{amsthm}
\usepackage{algorithmic}
\usepackage{graphicx}
\usepackage{makecell}
\usepackage{multirow}
\usepackage{textcomp}
\usepackage{float}
\usepackage{pifont}
\usepackage{hyperref}
\newcommand{\cmark}{\ding{51}}
\usepackage{xcolor}
\usepackage{booktabs}
\usepackage{booktabs}
\usepackage{siunitx}
\usepackage{colortbl}
\def\BibTeX{{\rm B\kern-.05em{\sc i\kern-.025em b}\kern-.08em
    T\kern-.1667em\lower.7ex\hbox{E}\kern-.125emX}}
\begin{document}

\newtheorem{proposition}{\textbf{Proposition}}

\newtheorem{theorem}{\textbf{Theorem}}

\newtheorem{corollary}{\textbf{Corollary}}

\newtheorem{remark}{\textbf{Remark}}

\title{ConDR: Conservative Reachability through Neural Hamilton–Jacobi Subsolutions}

\author{%
\IEEEauthorblockN{Kendra Givens\IEEEauthorrefmark{1},
Taylor T. Johnson\IEEEauthorrefmark{1},
Bryan C. Ward\IEEEauthorrefmark{1}, and
Thomas Beckers\IEEEauthorrefmark{2}}
\IEEEauthorblockA{\IEEEauthorrefmark{1}Computer Science, Vanderbilt University, United States \\
\{kendra.l.givens, taylor.johnson, bryan.ward\}@vanderbilt.edu}
\IEEEauthorblockA{\IEEEauthorrefmark{2}Institute of Measurement, Control and Microtechnology, Ulm University, Germany \\
thomas.beckers@uni-ulm.de}
}

\sisetup{
    separate-uncertainty = true,
    table-align-uncertainty = true
}

\maketitle

\begin{abstract}
Hamilton-Jacobi (HJ) Reachability identifies the unsafe states of a dynamical system from which it can be forced into failure under worst-case disturbance. However, classical grid-based solvers scale exponentially with state dimension. DeepReach improves scalability by replacing the grid with a neural value function trained on the HJ-Isaacs (HJI) variational inequality (VI), but penalizes both sides of the residual equally. This allows the learned value to lie above the true one and unsafe states to be misclassified as safe. We propose Conservative DeepReach (ConDR) which aims to learn a subsolution of the HJI VI such that the learned unsafe set contains the true one. ConDR enforces the obstacle branch and terminal condition by construction, pushes the HJI residual to a safety margin $\epsilon$ below zero with a one-sided loss, and prevents the degenerate solution of classifying everything as unsafe. We prove that the exact solution of the margin-shifted VI that ConDR trains toward is a subsolution of the original VI and its value lies within $\epsilon\tau$ of the true value function. We compare against DeepReach and MADR in the established systems of Air3D, Air6D, a 9D multi-vehicle system, and a 13D quadrotor. ConDR reports the lowest rate of unsafe state misclassified and the lowest closed-loop collision rates with only a tiny increase in intervention. On Air3D and Air6D, ConDR tolerates model mismatch that scales with $\epsilon$. At matched false-negative rates, ConDR's unsafe sets are smaller than the baselines showing that safety is not obtained by trivial over-approximation. 
\end{abstract}


\section{Introduction} 
As more complex and autonomous systems are deployed in safety-critical settings, verifying their behavior becomes increasingly important. These systems often have nonlinear dynamics, operate under control and state constraints, and may be affected by external or even adversarial disturbances. Hamilton-Jacobi (HJ) reachability provides worst-case safety guarantees under exactly these conditions. By solving for the value function, HJ reachability identifies the backward reachable tube (BRT), which represents the set of states from which the system can be forced into the target set, a chosen unsafe set of states, within the time horizon under worst-case disturbance. The value function can also be used to recover the optimal control policy.

However, HJ reachability suffers from the curse of dimensionality. Classical HJ solvers use level-set methods on a discretized grid \cite{HJ_Overview} where both memory and runtime grow exponentially with the dimension of the system state. Exact methods can extend beyond low-dimensional systems, but require the dynamics to decompose into self-contained subsystems \cite{decomp}. Recently, neural approaches have improved the scalability. DeepReach \cite{deepreach} overcomes this limitation by representing the value function by an implicit neural representation (INR) \cite{sitzmann2019siren} instead of a grid, which avoids direct dependence on spatial discretizations and makes higher-dimensional problems more tractable. The ExactBC \cite{singh2025exactimpositionsafetyboundary} variant removes the boundary condition loss by construction. MPC-guided Adversarial DeepReach (MADR) \cite{madr} expands on DeepReach and ExactBC by augmenting the loss with a self-generated signal from model predictive control (MPC) rollouts, which improves convergence. The learned value function is then used as a least-restrictive safety filter in closed loop rollouts \cite{clr}, \cite{hsu2023safetyfilterunifiedview}. Control barrier functions \cite{cbf} provide an alternative safety filter, but require the barrier function to be provided or learned beforehand, whereas HJ reachability only requires the dynamics.

Despite the scalability benefits, neural HJ approaches recover the exact value function only in the limit of a vanishing residual over the whole domain \cite{hofgard2024convergenceguaranteesneuralnetworkbased}. However, this is never achieved in practice, as the residual is typically nonzero at most points. As a result, the sign of the residual matters, i.e., a positive residual pushes the learned value function above the true one and results in higher false negatives (FNs) where unsafe states are classified as safe. A negative residual only enlarges the learned unsafe set resulting in higher false positives (FPs). Since both sides are penalized equally, FNs can be as frequent as FPs. Deploying a learned value function in a real control system requires some protection against the FNs. There exist methods that can bound the error of the model training \cite{lin2024verificationneuralreachabletubes}, \cite{li2025certifiablereachabilitylearningusing}, but they inflate the volume of the level-set by the size of the worst violation. Additionally, neither changes the training objective that produced the violation. 

We propose Conservative DeepReach (ConDR), a neural HJ reachability method that trains towards a subsolution of the HJI VI, such that the target BRT overapproximates the true one. We enforce the obstacle branch \cite{fisac2014reachavoidproblemstimevaryingdynamics} and terminal condition of the HJI VI at all times \emph{by construction}, using a one-sided L1 loss to push the residual to the safe side with safety margin $\epsilon$, and add a complementarity loss term to prevent the degenerate solution of declaring everything as unsafe. This results in smaller post-training corrections, which yield a tighter tube volume and fewer collisions during closed-loop safety-filtering rollouts, while tolerating model mismatch up to $\epsilon$. 

\noindent{\textbf{Contributions}}
Our key contributions are 
\begin{enumerate}
    \item ConDR: a conservative neural HJ reachability model that uses an easy to implement margin-shifted one-sided loss, trains toward a subsolution of the HJI VI, and composes with MADR.
    \item Guarantees that the BRT of the exact margin-shifted VI solution contains the true BRT, overapproximated by at most $\epsilon\tau$, and that a value function whose residual is at most $-\epsilon$ remains a subsolution under model mismatch up to $\epsilon$.
    \item Evaluation against DeepReach and MADR under matched experimental setups on Air3D, Air6D, a 9D multi-vehicle system, and a 13D quadrotor where ConDR has the lowest FN, subsolution violations, and collision rates in every setting tested.
    \item A tunable safety margin i.e. sweeping $\epsilon$ on Air3D and Air6D shows the tolerated model mismatch scales with $\epsilon$ with a linear accuracy cost in $\epsilon$.
\end{enumerate}
 
\section{Background}
\subsection{HJ Reachability}
\label{sec:hj}
Consider a dynamical system
\begin{equation}
    \label{EQ:system}
    \dot{x}(t) = f(x(t), u(t), d(t)),
\end{equation}
where $x(t) \in \mathcal{X} \subset \mathbb{R}^n$ is the system state, $u(t) \in \mathcal{U} \subset \mathbb{R}^m$ is the control input, and $d(t) \in \mathcal{D} \subset \mathbb{R}^k$ is the disturbance input. The dynamics are given by $f : \mathcal{X}  \times \mathcal{U} \times \mathcal{D} \to \mathbb{R}^n$, such that there always exists a unique solution for (\ref{EQ:system}). HJ reachability can be viewed as a two-player zero-sum differential game in which one player (disturbance) attempts to steer the system toward an unsafe set while the other (controller) attempts to steer it away within a time horizon $T \in \mathbb{R}_{> 0}$. Let the unsafe set $\mathcal{S}$ be represented implicitly by a Lipschitz continuous function $g : \mathbb{R}^n \to \mathbb{R}$, such that
\begin{equation}
\label{eq:unsafe}
    \mathcal{S} = \{x \in \mathbb{R}^n \mid g(x) \leq 0\},
\end{equation}
where $g(x)$ is non-positive inside the unsafe set and positive outside it. The cost is defined as
\begin{equation}
    J(x, u(\cdot), d(\cdot)) = \min_s g(x(s)), s \in [0, \tau], x(0)=x.
\end{equation}
where $\tau$ denotes the time remaining before the time horizon ends.
Assuming the controller maximizes the cost to keep the system away from $S$ and the disturbance minimizes it to push the system towards $S$, the value function $V\colon \mathbb{R}_{\geq 0}\times \mathbb{R}^n\to\mathbb{R}$ is
\begin{equation}
    V(\tau,x) = \sup_{u(\cdot)} \inf_{d(\cdot)} \; J(x, u(\cdot), d(\cdot)).
\end{equation}

The backward reachable tube (BRT) $\mathcal{B}$ is the set of states whose trajectories can reach the unsafe set within the remaining time $\tau$. It is given by the zero sub-level set of the value function
\begin{equation}
    \mathcal{B}(\tau) = \{x \in \mathbb{R}^n \mid V(\tau,x) \leq 0\}.
\end{equation}

The value function is the solution to the Hamilton--Jacobi--Isaacs (HJI) variational inequality (VI) given by
\begin{align}
            \max\{\partial_\tau \!V\! -\! H(x, \nabla_x\! V), \!V\!-\!g(x)\} &= 0 \label{EQ:HJIVI}\\
    V(0, x) &= g(x)\notag
\end{align}
with the Hamiltonian $H$ defined as
\begin{equation}
    H(x,p) = \max_{u \in \mathcal{U}} \; \min_{d \in \mathcal{D}} \; p^\top f(x,u,d),
\end{equation}
where $p = \nabla_x V(\tau,x)$.
We denote $\partial_\tau V - H(x, \nabla_x V)$ as the differential branch and $V-g(x)$ as the obstacle branch \cite{fisac2014reachavoidproblemstimevaryingdynamics}.

\subsection{DeepReach}
DeepReach \cite{deepreach} represents the value function with a SIREN, an implicit neural representation, and trains it in an unsupervised, physics-informed manner \cite{raissi2019physics} on the HJI VI.
DeepReach learns $V_{\theta}(\tau,x) \approx V(\tau,x)$ where $\theta$ denotes the network parameters, $\tau$ represents the time-to-go, and $x$ the spatial coordinates of the learned value function $V_\theta\colon \mathbb{R}_{\geq 0}\times \mathbb{R}^n\to\mathbb{R}$. The original DeepReach required an additional loss term to satisfy the terminal condition. The ExactBC \cite{singh2025exactimpositionsafetyboundary} variant removes that loss term and allows the model to satisfy the terminal condition by construction. It modifies the output of the neural network as
\begin{align}
    V_\theta(\tau, x) = g(x) + \tau \text{NN}_\theta(\tau, x)
\end{align}
such that when $\tau=0$, $V_\theta = g(x)$. The loss for ExactBC DeepReach is given by
\begin{align}
\mathcal{L}_\text{DR} = \mathbb{E}_{\tau, x}[|\max(\partial_\tau V_\theta - H(x, \nabla_x V_\theta), V_\theta - g)|],
\end{align}
which penalizes violations on both sides of the residual equally allowing unsafe states to be misclassified as safe. The model is trained with a curriculum over time-to-go. Spatial gradients of the value function needed to evaluate the Hamiltonian and recover the control policy, are obtained through automatic differentiation. We refer to ExactBC DeepReach as DeepReach throughout.

\subsection{MADR}
DeepReach's residual only signal can be too weak for high-dimensional systems so DeepReach-MPC \cite{feng2025bridgingmodelpredictivecontrol} added a supervised loss signal with the dataset obtained by model predictive control (MPC) rollouts. MPC-guided Adversarial DeepReach (MADR) \cite{madr} extends DeepReach MPC \cite{feng2025bridgingmodelpredictivecontrol} to handle adversarial input. It runs MPC across sampled states and times with the disturbance chosen adversarially from the gradient of the current value function. MADR uses the running-minimum cost as an approximate value function label and an L1 loss term is used to penalize the distance between the $V_\theta$ and the label. The labels are regenerated periodically from the model so no ground truth ever enters the model. An additional loss term is added that penalizes the network if $V_\theta$ calls a state safe that was not safe in the rollout. However, MADR still inherits the DeepReach problem of penalizing both sides of the residual equally. We also use the ExactBC version of MADR, referred to as just MADR throughout.

\subsection{Problem Statement}
Given a dynamical system (\ref{EQ:system}), an unsafe set of states $\mathcal{S}$ (\ref{eq:unsafe}) and a time horizon $T>0$, we aim to learn a neural value function $V_\theta$ approximating the largest subsolution of the margin-shifted HJI VI such that 
\begin{itemize}
    \item $V_\theta(\tau, x) \le V_{\text{true}}(\tau, x), \forall (\tau, x) \in [0, T] \times \mathcal{X}$, where $V_{\text{true}}$ represents the exact solution to the HJI VI (\ref{EQ:HJIVI})
    \item and the zero sub-level set $\{x\in\mathcal{X}: V_\theta(\tau, x) \le 0\}$ is an over-approximation containing the true Backward Reachable Tube $\mathcal{B}_{\text{true}}(\tau), \forall \tau \in [0, T]$.
\end{itemize}
At a high level, we aim to learn a value function $V_\theta$ that lies below the true value function $V$ so that its learned unsafe set over-approximates the true one to reduce the risk of misclassifying unsafe states as safe.

\section{Methods}
A subsolution for the HJI VI requires both the obstacle and differential branch in (\ref{EQ:HJIVI}) to be non-positive. We build on DeepReach to encourage $V_\theta$ to satisfy the subsolution property. ConDR constructs the output of the neural network to have a non-positive obstacle branch by construction and modifies the loss to train towards satisfying the differential branch without converging to the degenerate solution of declaring every state unsafe.

\subsection{HJI VI Subsolution}
The true HJI VI solution satisfies both the obstacle and differential branch inequalities and has at least one of the branches be an equality at every point. It additionally requires the terminal condition $V(0, x) = g(x)$. A valid subsolution drops the equality requirement i.e. both $V \le g$ and $\partial_\tau V - H(x, \nabla_x V) \le 0$ must be satisfied at every point without requiring the equality and requires a relaxed version of the terminal condition $V(0, x) \le g(x)$. In the following, we present new approaches for the obstacle and the differential branch towards this goal.

\subsection{Obstacle Branch}
We build on ExactBC's exact construction of the terminal condition to also enforce the obstacle branch $V \le g$. We constrain the correction to be nonpositive such that the exact construction then becomes
\begin{align}
    \label{EQ:exact}
    V_\theta(\tau, x) = g(x) - \tau\text{softplus}(\text{NN}_\theta(\tau, x)).
\end{align}
Therefore, the learned value function $V_\theta(\tau,x)$ is always less than or equal to $g(x)$ for every $\tau > 0$ and $x\in\mathcal{X}$ as shown later in Proposition \ref{prop:construction}. In this way, we can satisfy the inequality of the obstacle branch \emph{by construction}. 

\subsection{Differential Branch}
\label{sec:diff}
Unlike the obstacle branch, we cannot enforce the $V_\theta$ to be a valid subsolution of the differential branch through construction. Therefore, we train towards it through the loss term. 
For that, let $r = \partial_\tau V - H(x, \nabla_xV)$ be the residual and $r_\theta$ denote the residual with the learned $V_\theta$. Rather than target $r_\theta \le 0$, we instead introduce a safety margin $\epsilon > 0$ to target $r_\theta \le -\epsilon$. We define the margin-shifted VI, whose exact solution we denote $V_\epsilon$,
\begin{align}
    \label{EQ:marginVI}
    \max\{r_\epsilon + \epsilon, V_\epsilon-g(x)\} &= 0\\
    V_\epsilon(0, x) &= g(x)\notag.
\end{align}
ConDR trains $V_\theta$ toward $V_\epsilon$ using the losses below. To encourage $r_\theta \le -\epsilon$, we add the loss term 
\begin{align} \mathcal{L}_{\text{diff}} = \mathbb{E}[\text{relu}(r_\theta + \epsilon)].
\end{align}
If $r_\theta \le -\epsilon$, the loss term is 0. When $r_\theta > -\epsilon$, the loss term grows linearly with the error. Since the model can only sample a finite number of points and will always have some nonzero residual, $\epsilon$ allows the model to satisfy the original inequality despite approximation error i.e. a residual error smaller than $\epsilon$ still leaves $r_\theta \le 0$ and $V_\theta$ remains a subsolution to the original HJI VI. Setting $\epsilon = 0$ recovers a one sided version of the original DeepReach loss.

This loss term alone only penalizes residuals above $-\epsilon$ and is indifferent to how far below $-\epsilon$ they go. This can lead to a degenerate solution that predicts everything as unsafe. The original HJI VI prevents this since at least one of the branches must be an equality at each point. We prevent the degenerate solution by encouraging at least one of the branches to be tight i.e. to train towards the largest subsolution of the margin-shifted VI through the additional loss term 
\begin{align}
    \mathcal{L}_{\text{deg}} = \mathbb{E}[\min(|r_\theta+\epsilon|, g(x) - V_\theta)].
\end{align}
At each point, the loss equals the smaller of the two branches so that when one branch is tight, the loss is small.

We also add another loss term which penalizes residuals that exceed the margin on the safe side to aid in optimization. It is the negated version of the $\mathcal{L}_{\text{diff}}$ loss given by
\begin{align}
    \mathcal{L}_{\text{opt}} = \mathbb{E}[\text{relu}(-(r_\theta + \epsilon))].
\end{align}

The full objective is then 
\begin{align}
    \mathcal{L} = w_{\text{diff}}\mathcal{L}_{\text{diff}} + w_{\text{deg}}\mathcal{L}_{\text{deg}} + w_{\text{opt}}\mathcal{L}_{\text{opt}},
\end{align}
where $w_{\text{diff}},w_{\text{deg}},w_{\text{opt}} \in \mathbb{R} > 0$ are hyperparameters that weight each loss term.

\subsection{Formal Guarantees}
We prove that ConDR satisfies the obstacle branch and terminal condition exactly for every network in Proposition~\ref{prop:construction}. In Theorem~\ref{thm:bound}, we prove the solution of the margin-shifted VI that ConDR trains toward is a subsolution of the original VI. Corollary \ref{corl:cert}, proves that we can shift the BRT after training to contain the true BRT. Finally, in Theorem \ref{thm:mm}, we show that ConDR can tolerate model mismatch up to $\epsilon$ between the true and modeled Hamiltonian.
 
\begin{proposition}[Exact Construction] 
\label{prop:construction}
For every $\theta\in \mathbb{R}^p$, where $p$ denotes the number of parameters in the neural network of $V_\theta$, the construction (\ref{EQ:exact}) satisfies $V_\theta(0, x) = g(x)$ and $V_\theta(\tau, x) < g(x)$ for all $x\in\mathbb{R}^n$ and $\tau > 0$.
\end{proposition} 
\begin{proof}
At $\tau=0$, the second term in (\ref{EQ:exact}) is zero so $V_\theta(0, x) = g(x)$. For $\tau>0$, the correction is $\tau$ times a positive quantity. Subtracting a positive number from $g$ can only decrease $V_\theta$. Therefore, $V_\theta(\tau, x)=g(x)$ at $\tau=0$ and $V_\theta(\tau, x) < g(x)$ for all $\tau > 0$.
\end{proof} 
Proposition~\ref{prop:construction} shows that regardless of the neural network's parametrization, the obstacle branch and terminal condition will always be satisfied.

\begin{theorem}[Bounding the Subsolution]
    \label{thm:bound}
    Let $V_\epsilon$ represent the exact solution to the margin shifted VI (\ref{EQ:marginVI}). Then, for all $(\tau, x) \in [0, T] \times \mathbb{R}^n, V_{\text{true}}(\tau, x) - \epsilon\tau \le V_\epsilon(\tau, x) \le V_{\text{true}}(\tau, x)$. Therefore, for every $\tau \in [0, T]$, $\{x \in \mathbb{R}^n: V_{\text{true}}(\tau, x) \le 0\} \subseteq \{x \in \mathbb{R}^n: V_\epsilon(\tau, x) \le 0\} \subseteq \{x \in \mathbb{R}^n: V_{\text{true}}(\tau, x) \le \epsilon\tau\}$.
\end{theorem} 
\begin{proof}
Let $r[V] = \partial_\tau V - H(x, \nabla_x V)$ and let $B = V_{\text{true}}(\tau, x) - \epsilon\tau$ so that $\partial_\tau B = \partial_\tau V_{\text{true}}(\tau, x) - \epsilon$. For the upper bound, $r[V_\epsilon] \le -\epsilon \le 0, V_\epsilon \le g$ and $V_\epsilon(0, x) = g$, therefore, $V_\epsilon$ is a subsolution to the original VI and lies below the true solution by the comparison principle \cite{bardi1997optimal}. Therefore, $V_\epsilon \le V_{\text{true}}$.
    For the lower bound, we will prove each branch and terminal condition. For the differential branch, $\nabla_xB = \nabla_xV_{\text{true}}$, so $H(x, \nabla_xB) = H(x, \nabla_x V_{\text{true}})$. Additionally, $\partial_\tau B = \partial_\tau V_{\text{true}} - \epsilon$. Therefore, $r[B] = r[V_{\text{true}}] -\epsilon$. Since $r[V_{\text{true}}] \le 0$, $r[B] \le -\epsilon$.
    For the obstacle branch, $B = V_{\text{true}} -\epsilon\tau \le V_{\text{true}}$ since $\epsilon\tau \ge 0$ and $V_{\text{true}} \le g$ since $V_{\text{true}}$ satisfies the obstacle branch (\ref{EQ:HJIVI}). Therefore, $B \le g$.
    For the terminal condition, $\tau = 0$ reduces the $\epsilon\tau$ term to zero. Therefore, $B(0, x) = V_{\text{true}}(0, x) = g$.
    As a consequence, $B$ is a subsolution of the margin shifted VI and lower bounds $V_\epsilon$ meaning that $V_{\text{true}}-\epsilon\tau \le V_\epsilon \le V_{\text{true}}$ and $V_\epsilon$ is at most $\epsilon\tau$ away from $V_{\text{true}}$.
\end{proof}
\begin{remark}
    The upper bound $V_\epsilon \le V_{\text{true}}$ follows because $V_\epsilon$ is a subsolution of the original VI and lies below the exact solution $V_{\text{true}}$. However, $V_\epsilon$ could be arbitrarily below $V_{\text{true}}$ and needs a valid lower bound to prevent the degenerate solution of declaring every state as unsafe. In general, $V_{\text{true}}$ is not a subsolution of the margin VI since $r[V_{\text{true}}] = 0$ whenever the differential branch is active and the margin VI requires $r \le -\epsilon$. Therefore, we shift $r[V_{\text{true}}]$ by $\epsilon$. We cannot add $\epsilon$ to $H(x, \nabla_xV)$ without changing $\nabla_xV$ so we lower $\partial_\tau V$ by $\epsilon$ instead.
\end{remark}
Theorem \ref{thm:bound} proves that the exact solution to the margin-shifted VI that ConDR trains toward, $V_\epsilon$, lies below $V_{\text{true}}$ and that its BRT never misclassifies an unsafe state. Additionally, the BRT of $V_\epsilon$ overapproximates the true BRT by at most $\epsilon\tau$. Theorem \ref{thm:bound} concerns the exact subsolution $V_\epsilon$ but, in practice, a trained neural network can only satisfy the residual approximately. Corollary \ref{corl:cert} converts its worst violation into a level-set shift to certify the resulting backward reachable tube. 
\begin{corollary}[Certification]
    \label{corl:cert}
    Let $V_\theta\in C^1$ with $V_\theta(0, x) = g$ and let $\delta \ge 0$. If $V_\theta(\tau, x) \le g$ and $r_\theta \le \delta$ everywhere, then $V_\theta(\tau, x) \le V_{\text{true}}(\tau, x) + \delta\tau \le V_{\text{true}}(\tau, x) + \delta T$ so for all $\tau \in [0, T]$, the true BRT is contained within the set $\{x \in \mathbb{R}^n: V_\theta(\tau, x) \le \delta T\}$.
\end{corollary}
\begin{proof}
    Let $W = V_\theta - \delta\tau$, then $\nabla_xW = \nabla_xV_\theta$ and $\partial_\tau W = \partial_\tau V_\theta - \delta$ so that $r[W] = r_\theta - \delta \le 0$. Since $\delta\tau \ge 0$, $W \le V_\theta(\tau, x) \le g$ and $W(0, x)=g(x)$. Therefore, $W$ is a subsolution of the original VI and $W \le V_{\text{true}}$ so $V_\theta(\tau, x) \le V_{\text{true}} + \delta\tau$. At any state with $V_{\text{true}} \le 0$, this gives $V_\theta(\tau, x) \le \delta\tau \le \delta T$.
\end{proof}
Corollary \ref{corl:cert} converts a global bound $\delta$ of the residual violation into a level-set shift where the shifted, learned BRT $\{x \in \mathbb{R}^n: V_\theta(\tau, x) \le \delta T\}$ contains the true one whenever $r_\theta \le \delta$ holds everywhere. However, this can only be a probabilistic estimate with finitely many samples. 

\begin{theorem}[Model Mismatch]
    \label{thm:mm}
    Let $V_\theta(\tau, x) \le g$ with $V_\theta(0, x)=g(x)$. Let $\hat{H}$ be the Hamiltonian of the modeled/perturbed dynamics.
    If $r_\theta \le -\epsilon$ under $\hat{H}$ and $|H(x, \nabla_xV_\theta) - \hat{H}(x, \nabla_xV_\theta)| \le \epsilon$ for all $(\tau, x) \in [0, T]\times\mathbb{R}^n$ then $\partial_\tau V_\theta - H(x, \nabla_xV_\theta) \le 0$ so $V_\theta$ is a subsolution of the true system's VI (\ref{EQ:HJIVI}) and  $V_\theta \le V_{\text{true}}$ by the comparison principle \cite{bardi1997optimal}. Therefore, the true system's BRT is contained in the learned BRT. 
\end{theorem}
\begin{proof}
The obstacle branch and terminal condition do not involve the dynamics and hold by hypothesis, for ConDR by construction (\ref{EQ:exact}), while the differential branch is satisfied because
    $\partial_\tau V_\theta(\tau, x) - H = (\partial_\tau V_\theta(\tau, x) - \hat{H}) + (\hat{H}-H) \le -\epsilon + \epsilon = 0$ so $V_\theta(\tau, x)$ is a subsolution of the true VI (\ref{EQ:HJIVI}).
\end{proof}
Theorem \ref{thm:mm} proves that a residual trained $\epsilon$ below zero remains a valid subsolution under Hamiltonian perturbations up to $\epsilon$.

\section{Experimental Setup}
We evaluate ConDR on four systems from the DeepReach and DeepReach-MPC benchmarks all with horizon $T=1$.
\subsection{Systems}
Air3D \cite{air3d} is a three dimensional system modeling a two-aircraft collision-avoidance scenario in relative coordinates, relative position and heading, with the evader as the controller and the pursuer as the disturbance. Each has a bounded turn rate and a collision radius of 0.25. We compute the ground truth from a level-set solver on an $81^3$ grid. Air6D is the same scenario in absolute coordinates with a six dimensional state, where the ground truth can be obtained from the three-dimensional game. Multi-vehicle 9D \cite{deepreach} is a non-adversarial nine dimensional scenario with three cooperating Dubins car vehicles trying to avoid pairwise collisions. Quadrotor 13D \cite{feng2025bridgingmodelpredictivecontrol}, the highest dimensional system we evaluate on, is the DeepReach-MPC benchmark that models a non-adversarial single thirteen dimensional quadrotor with position, unit quaternion, velocity, and angular rate with the goal of avoiding the cylinder obstacle. For the Multi-vehicle 9D and Quadrotor 13D systems, the high dimensionality prevents the computation of a ground-truth solution.

\subsection{Models}
We compare against the baselines DeepReach and MADR both with the ExactBC L1 loss variant.
We also introduce ConDR-MADR as a composition of ConDR and MADR which combines the MADR MPC training signal with our ConDR loss. Both baselines are run from the released MADR implementation \cite{madr}.

\subsection{Training}
All the models use a three-layer SIREN network with 512 neurons per layer, the Adam optimizer with a learning rate of $2\cdot 10^{-5}$, and the DeepReach's curriculum over time-to-go, with 65,536 points per iteration (65,000 for the quadrotor). Air3D and Air6D train for 15k epochs with a 2k-epoch pretraining phase at $\tau=0$ and a 10k curriculum stage, multi-vehicle 9D, for 60k epochs with a 50k curriculum, and quadrotor 13D for 104k epochs with a 1k pretraining phase. ConDR uses the safety margin $\epsilon =0.02$ on Air3D, Air6D, and multi-vehicle 9D and $\epsilon=0.085$ on the quadrotor. The loss weights are set to $\text{w}_{\text{diff}}=1, \text{w}_{\text{deg}}=1.2, \text{w}_{\text{opt}}=0.15$ for every experiment. The MPC loss label uses 500 states per refinement stage on Air3D, Air6D, and multi-vehicle 9D, and 5k on the quadrotor. Air3D is trained across 5 seeds while every other system is trained across 3 seeds. Every statistic reported is the mean $\pm$ the standard deviation (where shown) across these seeds. 

\section{Results}
\subsection{Ground Truth Comparisons}
We evaluate 395,678 Monte Carlo samples for Air6D and turn the absolute coordinates back into relative Air3D coordinates. All metrics are computed at $\tau=T$. We report the Intersection over Union (IoU) between the learned and ground truth BRT, where IoU measures the overlap ratio and is given by $
    \mathrm{IoU}(\tau)
    =
    \frac{
        |\tilde{\mathcal{B}}(\tau)\cap \mathcal{B}(\tau)|
    }{
        |\tilde{\mathcal{B}}(\tau)\cup \mathcal{B}(\tau)|,
    } $
where $\tilde{\mathcal{B}}(\tau)$ denotes the predicted BRT and $\mathcal{B}(\tau)$ denotes the ground-truth BRT. We also report the FN and FP rates, the fraction of states where $V_\theta \le V_{\text{true}}$, and the largest violation of the obstacle branch $V_\theta - g \le 0$.

\begin{figure}[t]
    \centering
\includegraphics[width=\linewidth]{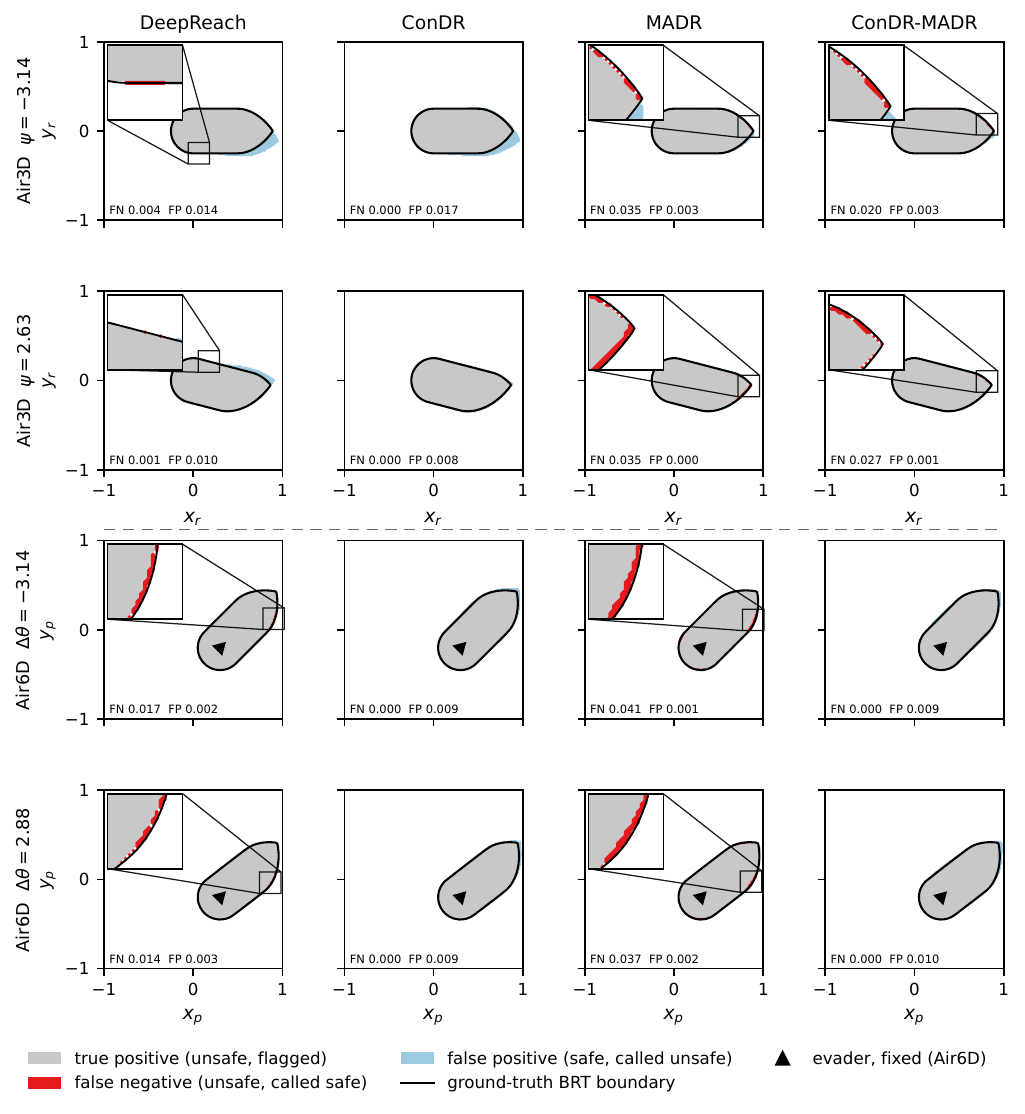}
    \vspace{-0.5cm}\caption{BRT classification at the slices where the baselines miss most. Top two rows: Air3D heading slices. Bottom two rows: Air6D pursuer-position slices with the evader fixed (triangle). Gray: true positives. Red: unsafe states called safe. Blue: safe states called unsafe. Black: true BRT boundary.\label{fig:air}}
\end{figure}

Table \ref{tab:gt} shows that across both systems, ConDR achieves the lowest FN and the highest FP of all models as an over-approximation should, since it prefers to classify safe states as unsafe rather than the reverse. On Air3D, it has zero FNs on four out of five seeds with one seed as an outlier. Additionally, ConDR satisfies the subsolution property $V_\theta \le V_{\text{true}}$ at 96\% of states on Air3D and 99\% on Air6D, against the baselines which show the wrong sign roughly half the time. ConDR also always satisfies the obstacle branch $V_\theta - g \le 0$, whereas DeepReach and MADR violate it. DeepReach shows a slightly higher IoU than ConDR because IoU counts misclassification of safe and unsafe states equally. ConDR-MADR beats MADR on nearly every metric, but satisfies $V_\theta \le V_{\text{true}}$ at a lower rate than ConDR due to the MPC loss pulling away from the pure residual loss. In Fig.~\ref{fig:air}, we show the BRTs at the headings where the baselines miss most. In both Air3D and Air6D, ConDR shows a clear over-approximation of the BRT, whereas the other models show misclassifications on the BRT boundary.
\begin{table*}[!t]
\centering
\caption{Ground-truth metrics at $\tau = T$ on Air3D (5 seeds) and Air6D (3 seeds), mean $\pm$ std. FN: unsafe states called safe. FP: safe states called unsafe. frac$(V_\theta \le V_{\text{true}})$: containment. $\max(V_\theta - g)$: worst obstacle-branch violation.}
\label{tab:gt}
\small
 \footnotesize
\setlength{\tabcolsep}{3pt}
\begin{tabular}{llccccc}
\toprule
System & Model & IoU $\uparrow$ & FN $\downarrow$ & FP $\downarrow$ & frac$(V_\theta \le V_{\text{true}})$ $\uparrow$ & $\max(V_\theta - g)$ $\downarrow$ \\
\midrule
\multirow{4}{*}{Air3D}
 & DeepReach  & $0.939 \pm 0.008$ & $0.0022 \pm 0.0004$ & $0.0059 \pm 0.0008$ & $0.659 \pm 0.016$ & $+0.018$ \\
 & \textbf{ConDR}      & $0.929 \pm 0.004$ & $\mathbf{0.0009 \pm 0.0018}$ & $0.0071 \pm 0.0006$ & $\mathbf{0.959 \pm 0.005}$ & $\mathbf{-0.005}$ \\
 & MADR       & $0.952 \pm 0.004$ & $0.0266 \pm 0.0035$ & $0.0021 \pm 0.0001$ & $0.440 \pm 0.043$ & $+0.027$ \\
 & \textbf{ConDR-MADR} & $\mathbf{0.963 \pm 0.002}$ & $0.0172 \pm 0.0013$ & $\mathbf{0.0020 \pm 0.0002}$ & $0.740 \pm 0.028$ & $\mathbf{-0.005}$ \\
\midrule
\multirow{4}{*}{Air6D}
 & DeepReach  & $\mathbf{0.950 \pm 0.002}$ & $0.0123 \pm 0.0024$ & $\mathbf{0.0032 \pm 0.0003}$ & $0.466 \pm 0.096$ & $+0.028$ \\
 & \textbf{ConDR}      & $0.924 \pm 0.004$ & $\mathbf{0.0006 \pm 0.0001}$ & $0.0065 \pm 0.0004$ & $\mathbf{0.994 \pm 0.001}$ & $\mathbf{-0.002}$ \\
 & MADR       & $0.933 \pm 0.006$ & $0.0196 \pm 0.0003$ & $0.0040 \pm 0.0005$ & $0.529 \pm 0.008$ & $+0.045$ \\
 & \textbf{ConDR-MADR} & $0.936 \pm 0.003$ & $0.0023 \pm 0.0005$ & $0.0052 \pm 0.0003$ & $0.930 \pm 0.001$ & $\mathbf{-0.004}$ \\
\bottomrule
\end{tabular}
\end{table*}













\subsection{Sampled Residual Bounds}
For systems without ground truth, we estimate a residual bound using the distribution-free conformal bound \cite{conform} that Lin and Bansal \cite{lin2024verificationneuralreachabletubes} apply to rollouts and instead apply it to the VI residual. We sample $n=200,000$ $(\tau, x)$ pairs uniformly from $[0, T]\times\mathcal{X}$ and compute the residual violation $v=\text{relu}(\max(r_\theta, V_\theta-g))$ which is zero if both inequalities are true. We set $\delta$ to be the violation at rank 198,137 so that it is slightly past the 99th percentile and with 0.999 confidence, this guarantees the violations are $\le \delta$ on 99\% of the sampling distribution. Although, this uses finite samples and does not hold globally as Corollary \ref{corl:cert} requires, we can still use Corollary \ref{corl:cert} to obtain a shifted tube $\{x\in\mathbb{R}^n: V_\theta \le \delta T\}$ rather than a formal certificate for ConDR and ConDR-MADR. For DeepReach and MADR, whose obstacle branch is not satisfied by construction, but bounded by $\delta$, we can use the same argument $W=V_\theta - \delta\tau - \delta$ to give the level $\delta(T+1)$.

Table \ref{tab:certificate} shows the fraction of samples where $\text{relu}(\max(r_\theta, V_\theta-g)) > 0$, the bound $\delta$, and the volume of the shifted BRT as a fraction of the state space. ConDR violates the subsolution property on 1\% to 3\% of sampled states against DeepReach and MADR who violate on 25\% to 52\% of states. ConDR also reports the smallest shifted-tube volume on every system. The certificate can also be checked against Air3D's ground truth where ConDR needs zero shift on four out of five seeds since its learned tube already covers 99\% of the true BRT without correction.

\begin{table}[t]
\centering
\caption{Sampled residual bound ($n = 200$K, $\eta = 0.01$, $\beta = 0.001$).
Violation: fraction of samples violating a subsolution inequality.
$\delta_{\text{sub}}$: bound on the violation holding on a $1-\eta$ fraction of the domain with confidence $1-\beta$.
Shifted vol.: fraction of the box in $\{V_\theta \le \delta T\}$ for ConDR and ConDR-MADR (exact obstacle branch) and $\{V_\theta \le \delta(T+1)\}$ for DeepReach and MADR (obstacle branch violated up to $\delta$).}
\label{tab:certificate}
\small
\footnotesize
\setlength{\tabcolsep}{3pt}
\begin{tabular}{llcccc}
\toprule
& & DeepReach & ConDR & MADR & \makecell{ConDR-\\MADR} \\
\midrule
\multirow{3}{*}{\rotatebox{90}{Air3D}}
& Violation $\downarrow$ & 0.25 & \textbf{0.01} & 0.49 & 0.11 \\
& $\delta_{\mathrm{sub}}$ $\downarrow$ & 0.018 & \textbf{0.004} & 0.056 & 0.096 \\
& Shift.\ vol.\ $\downarrow$ & 0.116 & \textbf{0.097} & 0.162 & 0.152 \\
\midrule
\multirow{3}{*}{\rotatebox{90}{Air6D}}
& Violation $\downarrow$ & 0.52 & \textbf{0.03} & 0.50 & 0.11 \\
& $\delta_{\mathrm{sub}}$ $\downarrow$ & 0.056 & \textbf{0.033} & 0.158 & 0.144 \\
& Shift.\ vol.\ $\downarrow$ & 0.129 & \textbf{0.094} & 0.251 & 0.149 \\
\midrule
\multirow{3}{*}{\rotatebox{90}{9D}}
& Violation $\downarrow$ & 0.49 & \textbf{0.03} & 0.48 & 0.07 \\
& $\delta_{\mathrm{sub}}$ $\downarrow$ & 0.083 & \textbf{0.054} & 0.128 & 0.109 \\
& Shift.\ vol.\ $\downarrow$ & 0.379 & \textbf{0.247} & 0.491 & 0.314 \\
\midrule
\multirow{3}{*}{\rotatebox{90}{13D}}
& Violation $\downarrow$ & 0.49 & \textbf{0.03} & 0.48 & 0.09 \\
& $\delta_{\mathrm{sub}}$ $\downarrow$ & 0.231 & \textbf{0.158} & 0.327 & 0.316 \\
& Shift.\ vol.\ $\downarrow$ & 0.124 & \textbf{0.071} & 0.170 & 0.097 \\
\bottomrule
\end{tabular}
\end{table}

\subsection{Closed-loop Safety Filters}
We deploy each learned value function as least-restrictive safety filters where the nominal controller acts unless $V_\theta \le 0$, in which case the optimal control extracted from $\nabla V_\theta$ overrides it. For Air3D, Air6D, and multi-vehicle 9D, the nominal controllers set their turn rate to zero and fly straight. For Quadrotor 13D, the nominal controller flies at the same initial velocity and orientation. The adversary for Air3D and Air6D takes two forms: pure pursuit, a static policy of orienting towards the evader at max turn rate, or as a learned adversary which is the worst-case disturbance computed from a fixed DeepReach model and applied to all models. Multi-vehicle 9D and Quadrotor 13D do not have adversaries. We use two starting state procedures. On Air3D and Air6D, where ground truth exists, we sample verified safe states as the start. On all four systems, we sample starting states at a distance from the unsafe set that are unverified as safe. For each procedure, we sample 400 starting states per seed, shared across each method and report the collision and intervention rate.

\begin{figure}[b]
    \centering
\includegraphics[width=\linewidth]{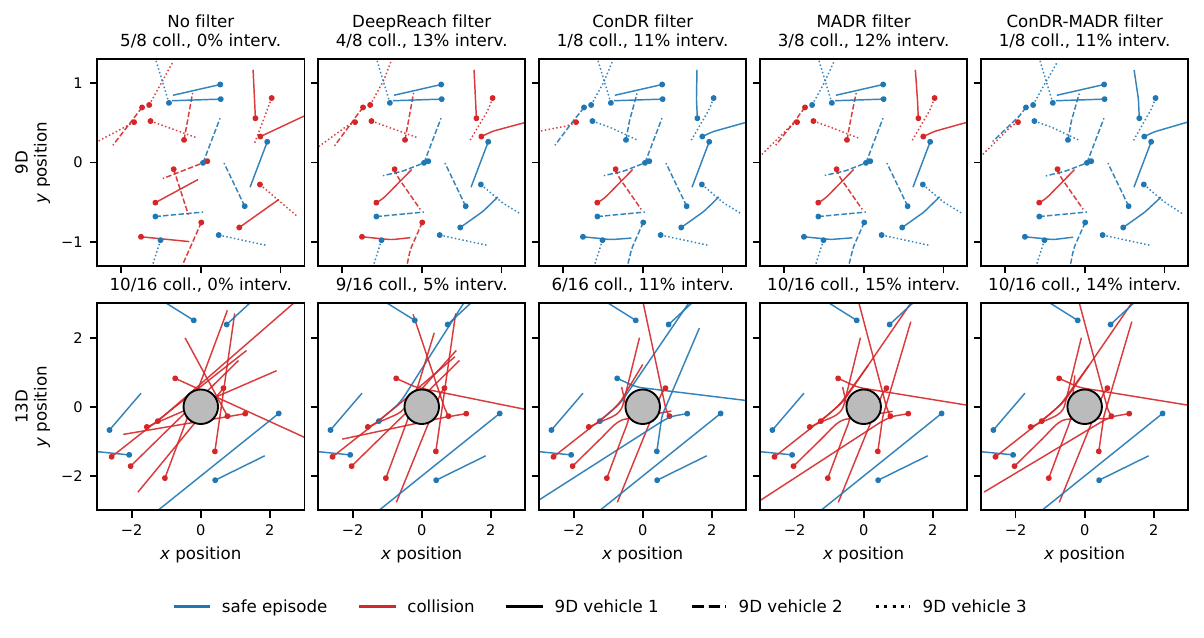}
    \caption{Rollouts from unverified starts: multi-vehicle 9D (8 starts), quadrotor 13D (16 starts). Red: collision}
    \label{fig:rollouts}
\end{figure}

In Table \ref{tab:filter}, ConDR achieves the lowest, sometimes zero, collision rate across all the models, while ConDR-MADR beats MADR in five cases, ties in one, and is only worse by 0.005 in the other two cases. Neither ConDR nor ConDR-MADR has to increase the intervention rate in order to lower their collision rate. In every column, they intervene at the same rate or by no more than 0.005 more with the exception of the quadrotor which intervenes on 1.4\% of the steps against DeepReach's 0.6\%.  In Fig. \ref{fig:rollouts}, we show example trajectories for each system from the unverified distance start state procedure. We choose starts where the nominal controller collides on most of the trajectories.
\begin{table}[t]
\centering
\caption{Closed-loop safety filtering: collision rate (mean $\pm$ std) / intervention fraction, 400 starts per seed. Top: verified-safe starts on Air3D and Air6D under pure-pursuit and learned adversaries. Bottom: unverified starts on all four systems. Bold: lowest collision rate.}
\label{tab:filter}
\small
\footnotesize
\setlength{\tabcolsep}{2pt}
\renewcommand{\arraystretch}{1.15}
\begin{tabular}{lcccc}
\toprule
\multicolumn{5}{l}{\textit{Verified-safe starts}} \\
& \multicolumn{2}{c}{Air3D} & \multicolumn{2}{c}{Air6D} \\
\cmidrule(lr){2-3} \cmidrule(lr){4-5}
Method & pursuit & learned & pursuit & learned \\
\midrule
No filter  & .178 & .215 & .140 & .160 \\
\arrayrulecolor{gray}\hline
DeepReach  & \makecell{.046$\pm$.005\\/.054} & \makecell{.040$\pm$.042\\/.074} & \makecell{.034$\pm$.008\\/.038} & \makecell{.098$\pm$.008\\/.060} \\
\arrayrulecolor{gray}\hline
ConDR      & \makecell{\textbf{.000$\pm$.000}\\/.055} & \makecell{\textbf{.004$\pm$.005}\\/.069} & \makecell{\textbf{.021$\pm$.005}\\/.037} & \makecell{\textbf{.078$\pm$.010}\\/.062} \\
\arrayrulecolor{gray}\hline
MADR       & \makecell{.090$\pm$.017\\/.053} & \makecell{.172$\pm$.010\\/.073} & \makecell{.048$\pm$.006\\/.037} & \makecell{.113$\pm$.015\\/.060} \\
\arrayrulecolor{gray}\hline
ConDR-MADR & \makecell{.077$\pm$.002\\/.058} & \makecell{.153$\pm$.018\\/.073} & \makecell{.043$\pm$.000\\/.038} & \makecell{.118$\pm$.006\\/.059} \\
\midrule
\multicolumn{5}{l}{\textit{Unverified starts}} \\
Method & Air3D & Air6D & 9D & 13D \\
\midrule
No filter  & .265 & .182 & .155 & .085 \\
\arrayrulecolor{gray}\hline
DeepReach  & \makecell{.083$\pm$.043\\/.094} & \makecell{.118$\pm$.008\\/.069} & \makecell{.126$\pm$.005\\/.052} & \makecell{.081$\pm$.003\\/.006} \\
\arrayrulecolor{gray}\hline
ConDR      & \makecell{\textbf{.047$\pm$.005}\\/.088} & \makecell{\textbf{.096$\pm$.010}\\/.071} & \makecell{\textbf{.042$\pm$.003}\\/.052} & \makecell{\textbf{.052$\pm$.007}\\/.014} \\
\arrayrulecolor{gray}\hline
MADR       & \makecell{.219$\pm$.014\\/.093} & \makecell{.132$\pm$.017\\/.070} & \makecell{.122$\pm$.003\\/.052} & \makecell{.079$\pm$.001\\/.012} \\
\arrayrulecolor{gray}\hline
ConDR-MADR & \makecell{.201$\pm$.018\\/.093} & \makecell{.137$\pm$.007\\/.067} & \makecell{.073$\pm$.012\\/.053} & \makecell{.079$\pm$.001\\/.013} \\
\bottomrule
\end{tabular}
\end{table}


\subsection{Model Mismatch}
We repeat the distance margin safety filter experiment with the dynamics perturbed to evaluate ConDR's robustness to model mismatch $M$. We scale the pursuer's speed by $1+M$ and the evader's by $1-M$, while the filter and learned adversary still use the nominal model. We retrain ConDR for each $\epsilon \in \{0, 0.02, 0.05, 0.10\}$ (three seeds each, five at $\epsilon=0.02$). 

\begin{figure}[t]
    \centering
\includegraphics[width=\linewidth]{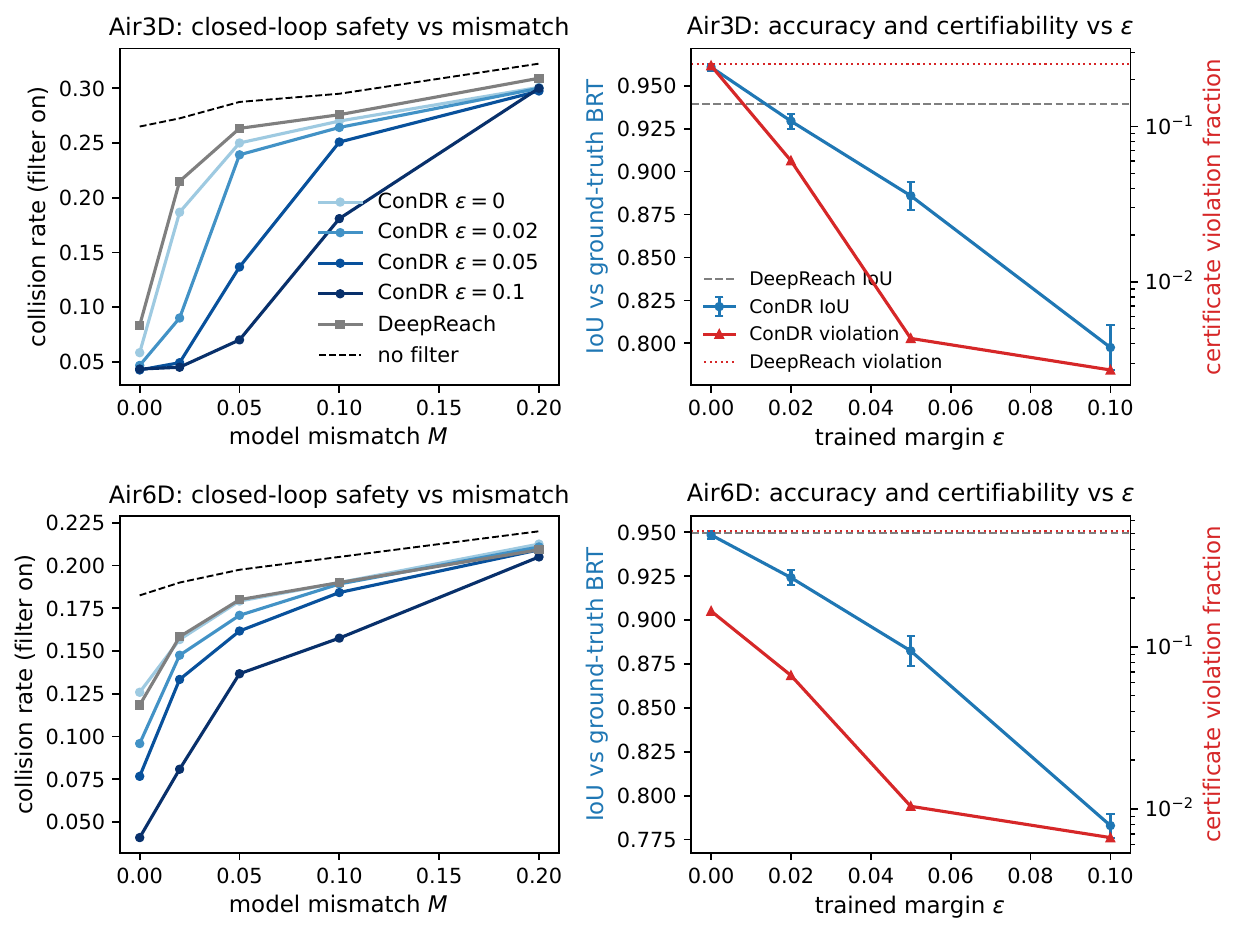}
    \caption{Air3D and Air6D. Left: collision rate versus mismatch $M$; each margin $\epsilon$ holds until $M \approx \epsilon$. Right: IoU (blue) and certificate violation (red, log) versus $\epsilon$.}
    \label{fig:margin_sweep}
\end{figure}

In Fig. \ref{fig:margin_sweep}, the left column shows that each model maintains its collision rate until the model mismatch roughly equals the safety margin $\epsilon$ and degrades after, showing consistency with Theorem \ref{thm:mm}. The $\epsilon=0$ model fails almost as fast as DeepReach. The right column shows that the IoU falls roughly linearly in $\epsilon$. The certificate violation fraction, where the model violates the subsolution inequality, shrinks as $\epsilon$ increases.

\subsection{Loss Ablation Study}
In Table \ref{tab:ablation}, we perform a loss ablation on ConDR and compare it to DeepReach. We add each loss component of ConDR to DeepReach one at a time and then remove them one at a time to test the effects of each term. The obstacle branch construction raises the subsolution property satisfaction from 66\% to 91\%, halves collisions, and improves IoU. As predicted by Section \ref{sec:diff}, without the $\mathcal{L}_\text{deg}$ loss, the model converges to the degenerate solution of classifying everything as unsafe. The $L_{\text{deg}}$ prevents the degenerate solution, while $L_{\text{opt}}$ increases the IoU and removes the last collisions. Adding the safety margin $\epsilon$ reduces the subsolution violations from 17\%-26\% to 0.4\%-1.2\%. Without $L_{\text{deg}}$, but with $\mathcal{L}_{\text{opt}}$ and the margin, the residual stays near the safety margin, but loses IoU.
\begin{table*}[!t]
\centering
\caption{Loss ablation on Air3D (5 seeds). Check marks show active components; rows without $\mathcal{L}_{\mathrm{diff}}$ use the two-sided residual. $^\dagger$Degenerate solution.}
\label{tab:ablation}
\small
\footnotesize
\setlength{\tabcolsep}{3pt}
\renewcommand{\arraystretch}{1.15}
\begin{tabular}{lccccc|cccccc}
\toprule
\textbf{Model}
& \makecell{\textbf{Obstacle}\\\textbf{construction}}
& $\mathcal{L}_{\mathrm{diff}}$
& $\mathcal{L}_{\mathrm{deg}}$
& $\mathcal{L}_{\mathrm{opt}}$
& $\varepsilon$
& IoU $\uparrow$
& FN $\downarrow$
& FP $\downarrow$
& frac$(V_\theta \le V_{\text{true}})$ $\uparrow$
& \makecell{Subsol.\\viol.\ $\downarrow$}
& Collisions $\downarrow$ \\
\midrule
DeepReach  & -- & -- & -- & -- & -- & 0.939 & 0.0022 & 0.0059 & 0.659 & 0.253 & 0.046 $\pm$ 0.005 \\
           & \cmark & -- & -- & -- & -- & \textbf{0.963} & 0.0014 & \textbf{0.0035} & 0.907 & 0.234 & 0.019 $\pm$ 0.006 \\
$\dagger$  & \cmark & \cmark & -- & -- & $0$ & 0.086 & 0.0000 & 1.0000 & 1.000 & 0.000 & 0.348 $\pm$ 0.024 \\
           & \cmark & \cmark & \cmark & -- & $0$ & 0.919 & \textbf{0.0002} & 0.0083 & 0.924 & 0.170 & 0.022 $\pm$ 0.005 \\
           & \cmark & \cmark & \cmark & \cmark & $0$ & 0.962 & 0.0021 & \textbf{0.0035} & 0.937 & 0.256 & 0.013 $\pm$ 0.005 \\
           & \cmark & \cmark & \cmark & -- & $0.02$ & 0.898 & 0.0014 & 0.0106 & 0.949 & \textbf{0.004} & 0.004 $\pm$ 0.003 \\
\midrule
           & \cmark & \cmark & -- & \cmark & $0.02$ & 0.707 & 0.0006 & 0.0392 & \textbf{0.977} & 0.006 & \textbf{0.000 $\pm$ 0.000} \\
           & -- & \cmark & \cmark & \cmark & $0.02$ & 0.924 & 0.0107 & 0.0068 & 0.731 & 0.172 & 0.090 $\pm$ 0.019 \\
\midrule
ConDR      & \cmark & \cmark & \cmark & \cmark & $0.02$ & 0.929 & 0.0009 & 0.0071 & 0.959 & 0.012 & \textbf{0.000 $\pm$ 0.000} \\
\bottomrule
\end{tabular}
\end{table*}

\subsection{Volume at Matched Safety}
To evaluate whether ConDR's low FN rate is simply a consequence of trivially overapproximating the learned BRT, we slide the level-set threshold $c$ and take the resulting BRT $\{x\in\mathbb{R}^n: V_\theta \le c\}$ then measure the volume of each model where volume is the fraction of the state space occupied by the learned BRT. If ConDR and ConDR-MADR were just DeepReach with a conservative offset, then their volumes would be equal at matched FNs. However, Table \ref{tab:matched_fn} shows that for Air3D, ConDR's and ConDR-MADR's volume is smaller than DeepReach/MADR and on Air6D is within 0.0003. Therefore, our method's conservatism is a result of where the loss places the BRT boundary and not simply from over-approximation.

\begin{figure}[b]
    \centering
\includegraphics[width=\linewidth]{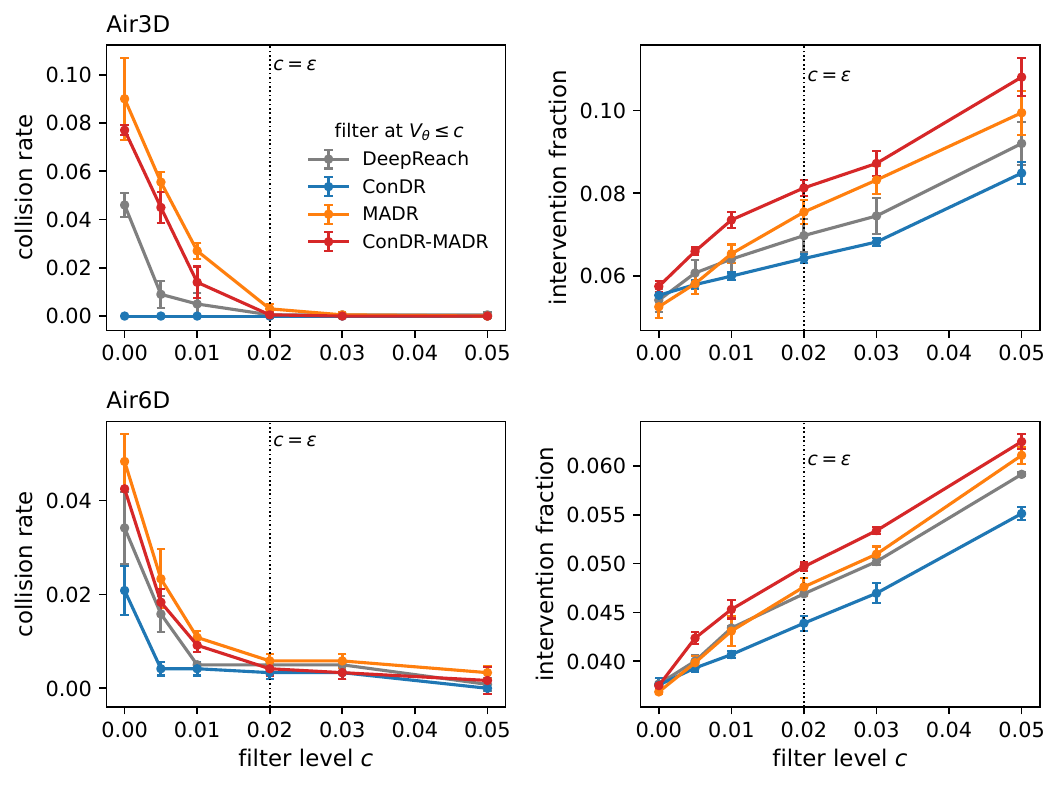}
    \caption{Collision rate (left) and intervention (right) as the filter level $c$ is raised from 0 on verified-safe starts; dotted line at $c = \epsilon$.}    \label{fig:shifted_threshold}
\end{figure}

We also evaluate the effect of sliding $c$ in the closed-loop safety filter experiment with verified safe starts. We measure the resulting collision and intervention rates for each model in Air3D and Air6D. Fig. \ref{fig:shifted_threshold}
shows that ConDR has the fewest collisions at every $c$ on both systems with equal or lower intervention. On Air3D, ConDR has a zero-collision rate, whereas the baselines only achieve this with $c\ge 0.03$ and must intervene 60\%-80\% more than ConDR at $c=0$. Therefore, shifting DeepReach/MADR after training can achieve ConDR's level of safety but only at the cost of more interventions. 
\begin{table}[t]
\centering
\caption{Volume of $\{V_\theta \le c\}$ after setting $c$ to the smallest value meeting each FN target, mean $\pm$ std over seeds (Air3D: 5, Air6D: 3). Bold: smallest in column.}
\label{tab:matched_fn}
\small
\footnotesize
\setlength{\tabcolsep}{2pt}
\renewcommand{\arraystretch}{1.0}
\begin{tabular}{@{}lccc@{}}
\toprule
Model & FN $\leq$ 0.05 & FN $\leq$ 0.03 & FN $\leq$ 0.01 \\
\midrule
\multicolumn{4}{@{}l}{Air3D (exact BRT volume 0.0863)} \\
DeepReach  & $0.0854\!\pm\!0.0008$ & $0.0876\!\pm\!0.0008$ & $0.0900\!\pm\!0.0008$ \\
ConDR      & $0.0827\!\pm\!0.0003$ & $0.0850\!\pm\!0.0003$ & $\mathbf{0.0882\!\pm\!0.0003}$ \\
MADR       & $0.0835\!\pm\!0.0001$ & $0.0856\!\pm\!0.0002$ & $0.0886\!\pm\!0.0003$ \\
ConDR-MADR & $\mathbf{0.0822\!\pm\!0.0001}$ & $\mathbf{0.0844\!\pm\!0.0002}$ & $0.0884\!\pm\!0.0002$ \\
\midrule
\multicolumn{4}{@{}l}{Air6D (exact BRT volume 0.0736)} \\
DeepReach  & $0.0716\!\pm\!0.0001$ & $0.0734\!\pm\!0.0002$ & $0.0760\!\pm\!0.0002$ \\
ConDR      & $0.0716\!\pm\!0.0003$ & $0.0736\!\pm\!0.0003$ & $0.0763\!\pm\!0.0004$ \\
MADR       & $0.0720\!\pm\!0.0004$ & $0.0743\!\pm\!0.0004$ & $0.0779\!\pm\!0.0006$ \\
ConDR-MADR & $\mathbf{0.0711\!\pm\!0.0001}$ & $\mathbf{0.0730\!\pm\!0.0002}$ & $\mathbf{0.0754\!\pm\!0.0002}$ \\
\bottomrule
\end{tabular}
\end{table}

\section{Conclusion}
We introduced ConDR that trains toward the solution of the margin-shifted VI. The solution itself is a subsolution of the original VI so its  BRT contains the true BRT and is within $\epsilon\tau$ of the true value function. Across Air3D, Air6D, multi-vehicle 9D, and quadrotor 13D, ConDR achieves the lowest FN rates, the lowest closed-loop collision rates, and the fewest subsolution violations, all while having the tightest shifted BRT volume. Composing with MADR improves MADR's FN, violation fraction, and shifted tube volume. It also lowers its closed-loop collision rates in five out of eight settings. While ConDR can only approximate the exact solution of the margin-shifted VI, the guarantees of Theorem \ref{thm:bound} hold for the exact solution. Finally, ConDR is especially beneficial for higher dimensional systems whose ground truth cannot be computed and is simple to integrate into existing DeepReach applications.


\bibliographystyle{IEEEtran}
\bibliography{bib}

\end{document}